\documentclass[11pt]{article}

\usepackage[final]{acl}

\usepackage{times}
\usepackage{latexsym}

\usepackage[T1]{fontenc}
\usepackage[utf8]{inputenc}

\usepackage{microtype}

\usepackage{inconsolata}

\usepackage{graphicx}
\usepackage[dvipsnames]{xcolor}
\usepackage{tcolorbox}

\usepackage{booktabs}
\usepackage{tabularx}
\usepackage{xcolor}
\usepackage{caption}
\usepackage{amsmath}
\usepackage{subcaption}
\usepackage{multirow}
\usepackage{cleveref}
\usepackage{enumitem}

\usepackage{algorithm}
\usepackage{algpseudocode}
\usepackage{ccg}

\definecolor{highlight}{RGB}{0, 102, 204}
\definecolor{derivation}{gray}{0.3}
\definecolor{highlightbg}{RGB}{255, 255, 153} % soft yellow
\definecolor{typegray}{gray}{0.3}

\setlist[itemize]{itemsep=2pt,topsep=2pt,parsep=0pt}

\newcommand{\hidden}[1]{}

\Crefname{section}{\S}{\S\S}

\usepackage{amsmath,amsfonts,bm,mathtools,amssymb}
\usepackage{tikz-dependency}
\usepackage{adjustbox}

\title{Typological Alignment of Stack-Based Language Models \\ on Mildly Context-Sensitive Artificial Languages}

\author{Nadine El-Naggar$^{1}$  \quad
  Tatsuki Kuribayashi$^{1,2}$ \quad 
  Ted Briscoe$^{1}$ \\
  $^1$Mohamed bin Zayed University of Artificial Intelligence \quad
  $^2$Tohoku University \\
    \texttt{\{nadine.naggar, tatsuki.kuribayashi, ted.briscoe\}@mbzuai.ac.ae} \\
}

\begin{document}
\maketitle
\begin{abstract}

Some properties of languages, e.g., subject-object-verb (SOV) word order, are more prevalent than others among the thousands of attested natural languages (NLs).
Such typological commonality is often attributed to learning biases. Computational simulations, recently with language models (LMs), have facilitated the exploration of this theory.
In this paper, we extend existing analyses of the relationship between LMs' learning biases and typological commonality on both data and model sides, focusing on: (i) cross-serial dependencies, the upper limit of attested syntactic complexity, and (ii) stack-based LMs (SLMs), potentially facilitating learning of hierarchical patterns. 
We first evaluate generalization of SLMs on cross-serial dependencies across diverse artificial languages and confirm that they struggle with such constructions. 
However, SLMs with limited working memory generalize better suggesting a possible basis for such inductive bias and thus the typological commonality of some word order configurations.

\end{abstract}

\section{Introduction}

Attested natural languages (NLs) exhibit both typological commonalities and considerable diversity. NLs are widely believed to be mildly context-sensitive~\cite{shieber1985evidence,li2024transformer}, yet they vary along numerous dimensions, including word order.
Such features are not randomly distributed; for example, Subject-Object-Verb (SOV) base word order is substantially more common than Object-Subject-Verb (OSV) order, reflecting skewed typological tendencies.

Linguists have argued that such skewed typological distributions could be attributed to learning bias~\cite{Kirby2008-zg,Culbertson2012-wc}: grammars that are easier to learn are more likely to survive successive (re-)learning.
Although learning bias here refers to human learners, language models (LMs) have been used to evaluate and refine this theory.
Recent work, for example, has used LMs, as a tool, to isolate the contribution of domain-general learning biases in simulations of typological patterns~\cite{xu2025can}. 
More broadly, LMs and computational simulations enable controlled manipulations of language learning conditions that would be ethically or practically infeasible in human experiments~\cite{warstadt2022what}, thereby providing a proof-of-concept for linguistic hypotheses that have not been quantitatively tested
~\cite{DBLP:conf/acl/KalliniPFMP24,DBLP:conf/acl/KuribayashiUYOB24,el2025word,yang-etal-2025-anything,hunter2025kallini}.

Most LM-driven research on typological word-order distributions, like \citet{el2025word} and \citet{DBLP:conf/acl/KuribayashiUYOB24}, has used artificial languages (ALs).
Such ALs are simplified \cite{DBLP:conf/coling/SomeyaYO24} and lack the full  attested complexity of NLs \cite{el-naggar-etal-2025gcg,xu2025can}, such as cross-serial dependency (CSD) constructions.
CSDs are mildly context-sensitive constructions attested in some NLs \cite{shieber1985evidence,maclachlan2002cross} and generally taken to reflect the upper bound of syntactic complexity.
CSDs are exemplified with `Swiss-German' sentences, such as \textit{We \textcolor{red}{Hans} \textcolor{blue}{the house} \textcolor{red}{helped} \textcolor{blue}{paint}}, where the dependencies cross (as indicated by coloring).

Certain stack-based language models (SLMs) are theoretically capable of learning some mildly context-sensitive languages and all context-free grammars~\cite{DBLP:conf/iclr/DuSell023}, roughly matching the structural complexity observed in NLs.
SLMs have also been shown to exhibit greater data efficiency~\cite{dusell2024stack} and hierarchical generalization~\cite{dusell2025bearingsyntacticfruitstackaugmented}.
But this research does not address whether SLMs can empirically generalize to CSDs in more realistic ALs rather than on a binary mirror language \{$ww'$\ |\ $w\in \{0,1\}^*$\}~\cite{DBLP:conf/iclr/DuSell023}, and, if so, what detailed inductive biases they exhibit.

In this study, we extend existing LM-driven research on typological word-order distributions \cite{DBLP:conf/acl/KuribayashiUYOB24,el2025word} to address three main research questions: first, how do different LM architectures and stack representations affect LM generalization to increasingly complex cross-serial dependencies (CSD) constructions? Second, how does the interaction between the underlying LM architecture and stack representation affect generalization?
Finally, how do the learning biases of these models over word-order configurations align with the typological distributions observed for natural languages?

\section{Background}

\subsection{Artificial Languages}

NLs differ from each other in many aspects, which makes it difficult to isolate specific linguistic properties by just comparing the learnability of different NLs.
As an alternative approach, artificial languages (ALs) are commonly used to simulate specific conditions for training and evaluating language learners, including both humans~\cite{Kirby2008-zg} and computational models~\cite{Wang2016-md,DBLP:conf/iclr/DeletangRGGWCCH23,DBLP:conf/coling/SomeyaYO24,el-naggar-etal-2025gcg}.
As discussed in~\citet{xu2025can}, there are mainly two approaches to create ALs: (i) creating a counterfactual version of the NL corpus by applying modification rules, or (ii) developing a grammar to generate a corpus from scratch. 
Both approaches have pros and cons.
The former will maintain rich and realistic properties in the NL corpus, but may retain errors and biases from rule-based modifications as well as many heterogeneous properties of the original NLs~\cite{hunter2025kallini}.
On the other hand, grammar-based corpus generation supports better control of the data, but potentially lacks realistic NL properties depending on the quality and comprehensiveness of the predefined grammar.
This study uses grammar-based AL generation but adds novel NL-like syntactic complexity, namely, various constructions with CSDs, into the training and test data.

\subsection{Typological Alignment of LMs}
What universal properties do NLs have, and where do they come from?
This is a fundamental question 
in linguistics and cognitive science \cite{greenberg1963some,comrie1989language}.
Recently, computational linguists have revisited this question using LMs \cite{linzen2016assessing,hupkes2020compositionality}.
This revival was partly triggered by claims that the unlimited capacity of (Transformer) LMs makes them irrelevant to the scientific pursuit of identifying the boundary between possible and impossible language \cite{marcus2018deep,bender2020climbing}.
In spite of this, empirical investigations show that LMs' learning biases typically align with typologically plausible patterns, highlighting the potential role of domain-general learning biases for explaining typological commonality \cite{xu2025can}.
Recently, this research has been expanded using quantitative typological databases, e.g. WALS \cite{wals}, and the correlation between LMs' performance and typological commonality has been studied more comprehensively.
Our study is also in line with such endeavors by extending two dimensions of AL complexity as well as testing LM variant models.

\subsection{Stack-Based LMs (SLMs)}

Given the claim that human language generalization and processing are biased toward hierarchical structure \cite{Chomsky1957-CHOSS-2,zaccarella2017neurobiological}, injecting structural bias into neural LMs has been explored.
Supervised syntactic LMs, such as recurrent neural network grammars (RNNGs)~\cite{dyer2016recurrent}, have been compared to base LMs~\cite{Hale2018FindingSearch,DBLP:conf/acl/KuribayashiUYOB24}, but the provision of syntactic annotation during training confounds the effect of supervision with model architecture.

SLMs are one variant, where stack-based memory is implemented as a differentiable sub-module, emulating a push-down automaton (PDA) facilitating the processing of hierarchically nested structures characteristic of context-free languages (CFLs).
Following~\citet{dusell2025bearingsyntacticfruitstackaugmented}, we test two SLM variants with superposition stacks (SUP)~\cite{joulin2015inferring} and nondeterministic stacks (ND)~\cite{dusell2024stack} using a standard LM self-supervised training regime. 

SLMs have a stack-based sub-layer, which updates the input representation $\bm x_t$ into $\bm x'_t$ using stack $S_t$ that maintains internal state using a stack data structure:
\begin{align}
    \bm a_t, \bm v_t &= \mathrm{Action}(\bm x_t) \;\;\mathrm{,}\\
    S_t &= \mathrm{Stack}(S_{t-1}, \bm a_t, \bm v_t) \;\;\mathrm{,} \\
    \bm r_t &= \mathrm{Read}(S_t) \;\;\mathrm{,} \\
    \bm x_t' &= f(\bm r_t, \bm x_t) \;\;\mathrm{.}
\end{align}
\noindent
Specifically, the stack is manipulated through actions (e.g, push or pop) weighted by $\bm a_t$ and a new item $\bm v_t$ to be pushed to the stack,  given the prediction $\mathrm{Stack(\cdot)}$ based on current input $\bm x_t$. 
A top-item $\bm r_t$ of the stack is then read by $\mathrm{Read}(\cdot)$, and the representation is updated to be $\bm x'_t$ with $\bm r_t$, ultimately predicting the next word.
Plausible actions are not limited to one at each processing step in order to handle (local) ambiguity.
ND SLMs store a set of different possible action histories and can refer to them as if they had many separate \textit{tapes} to record them.
In contrast, SUP SLMs superimpose the results of different stack actions and softly write them onto only one single tape.
As a result, ND SLMs can theoretically recognize all the CFLs, but there is no such guarantee for SUP SLMs.
See~\citet{dusell2024stack} for a more formal description and~\citet{dusell2025bearingsyntacticfruitstackaugmented} for the exact implementations we follow.

Recent studies explore SLMs' learning of formal ALs~\cite{DBLP:conf/iclr/DuSell023,dusell2024stack}, NLs~\cite{dusell2024stack}, and hierarchical generalization~\cite{dusell2025bearingsyntacticfruitstackaugmented}.
Still, detailed evaluation has been limited to very simplified ALs, such as palindromes \{$ww^R\ |\ w\in \{0,1\}^* $\} and has not explored the typological alignment of such models.

\begin{table*}[t]
\centering
\small
\begin{tabular}{ll}
\toprule
GCG Lexical Syntactic Category \cite{el-naggar-etal-2025gcg,el2025word} & Example  \\
\cmidrule(lr){1-1} \cmidrule(lr){2-2}

\textbf{Noun Phrase (NP)} -- {\color{typegray}NP} 
& \textbf{Tom} ga chased \textbf{Jerry} o \\

\textbf{Subject Marker} -- {\color{typegray}NP$_\text{SUBJ}$$\backslash$NP} 
& Tom \textbf{ga} chased Jerry o \\

\textbf{Object Marker} -- {\color{typegray}NP$_\text{OBJ}$$\backslash$NP} 
& Tom ga chased Jerry \textbf{o} \\

\textbf{Transitive Verb (VT)} -- {\color{typegray}(S$\mid_1$NP$_\text{SUBJ}$)$\mid_1$NP$_\text{OBJ}$} 
& Tom ga \textbf{chased} Jerry o \\

\textbf{Intransitive Verb (VI)} -- {\color{typegray}S$\mid_1$NP$_\text{SUBJ}$} 
& grey cat ga \textbf{fell} \\

\textbf{Verb with Complement (VCOMP)} -- {\color{typegray}(S$\mid_1$NP$_\text{SUBJ}$)$\mid_1$SCOMP} 
& Kim ga \textbf{believed} that Sandy o lied \\

\textbf{Complementizer (COMP)} -- {\color{typegray}SCOMP$\mid_2$S} 
& Kim ga believed \textbf{that} Sandy o lied \\

\textbf{Preposition (PREP)} -- {\color{typegray}(NP$\mid_3$NP)$\mid_3$NP} 
& elf \textbf{on} shelf ga laughed \\

\textbf{Adjective (ADJ)} -- {\color{typegray}NP$|_4$NP} 
&\textbf{grey} cat ga fell \\

\textbf{Relativizer (REL)} -- {\color{typegray}(NP$_\text{SUBJ}$$\mid_5$NP$_\text{SUBJ}$)$\mid_5$(S$\mid_1$NP$_\text{OBJ}$)} 
& man ga \textbf{whom} I ga met laughed \\

\textbf{Conjunction} -- {\color{typegray}Var$\backslash$Var$/$Var} 
& Tom \textbf{and} Jerry ga caused trouble  \\
\midrule
New GCG Lexical Syntactic Category & Example  \\
\cmidrule(lr){1-1} \cmidrule(lr){2-2}

\textbf{Dative Object Marker} -- {\color{typegray}NP$_\text{NI}$$\backslash$NP} & Tom ga helped Jerry \textbf{ni} cause trouble o\\

\textbf{Acc. Causative Verb (V1)} - {\color{typegray}((S$\mid_1$NP$_\text{SUBJ}$)$\mid_1$NP$_\text{OBJ}$)$\mid_1$(S$\mid_1$NP$_\text{SUBJ}$)} 
& I ga \textbf{let} children o help Kim ni paint house o \\

\textbf{Dat. Causative Verb (V2)} - {\color{typegray}((S$\mid_1$NPS$_\text{SUBJ}$)$\mid_1$NP$_\text{NI}$)$\mid_1$(S$\mid_1$NP$_\text{SUBJ}$)} 
& I ga let children o \textbf{help} Kim ni paint house o \\

\textbf{Object Relativizer (RELOBJ)} -- {\color{typegray}(NP$_\text{OBJ}$$\mid_5$NP$_\text{OBJ}$)$\mid_5$(S$\mid_1$NP$_\text{OBJ}$)} 
& I ga saw man o \textbf{who} I ga met\\

\textbf{Dative Obj. Relativizer (RELNI)} -- {\color{typegray}(NP$_\text{NI}$$\mid_5$NP$_\text{NI}$)$\mid_5$(S$\mid_1$NP$_\text{OBJ}$)} 
& I ga drove car ni \textbf{which} I ga liked\\

\bottomrule
\end{tabular}
\caption{
Lexical syntactic categories, their derivations, and their examples, where a word corresponding to the category is in bold.
The top half shows the categories defined by \citet{el-naggar-etal-2025gcg, el2025word}. 
The bottom half shows the new categories that we added to extend the grammar.
The vertical bars ``$\mid$'' in the GCG lexical syntactic categories represent either forward or backward slashes, and those with the same index are controlled by the same word order parameter (see Table~\ref{tab:wo_params}).
}
\label{table:category_descriptions}
\end{table*}

\begin{table}[hbt]
    \centering
    \small
    \begin{tabular}{lp{2.2cm}p{2.2cm}}
    \toprule
        Digit & 0 & 1 \\ 
        \midrule
        1 (Base) & SOV & VOS \\
        2 (COMP) & Preposed complementizer & Postposed complementizer \\
        3 (PP) & Postposition & Preposition \\
        4 (ADJ) & Prenominal adjective & Postnominal adjective \\
        5 (REL) & Preposed relativizer & Postposed relativizer \\
        \bottomrule
    \end{tabular}
    \caption{Word order parameters and the constructions introduced by their assignment (0/1). The exact implementation is shown in Table~\ref{table:category_descriptions}.}
    \label{tab:wo_params}
\end{table}

\begin{figure*}[t]
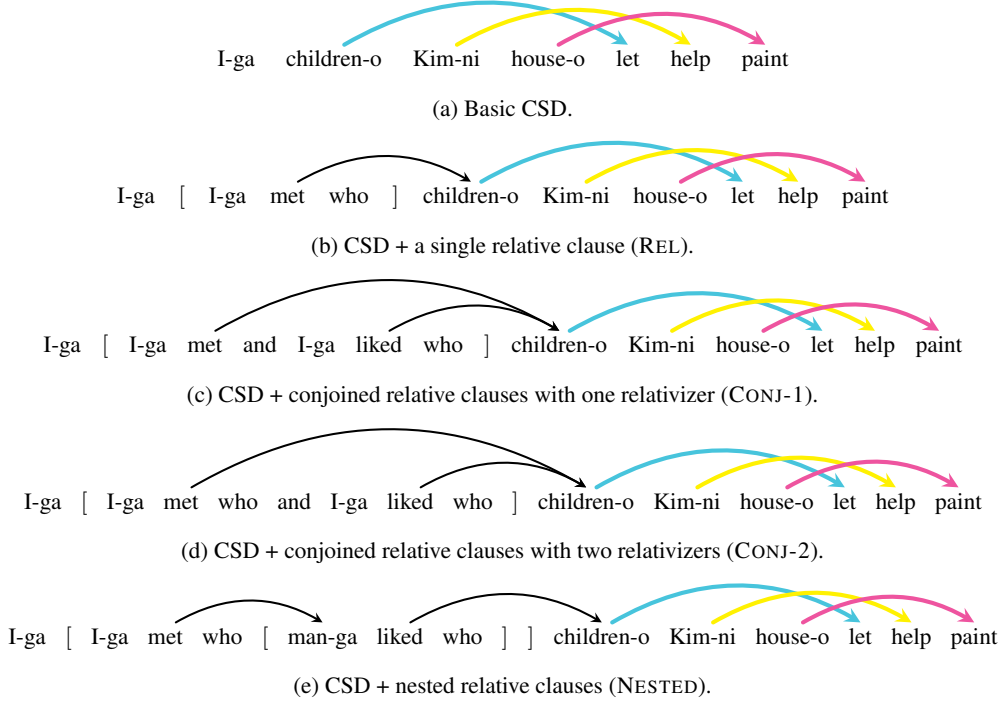

\centering

\begin{adjustbox}{max width=\textwidth}
\begin{tabular}{c}

%================================================
% Basic
%================================================
\begin{subfigure}[t]{\textwidth}
\centering
\small
\begin{dependency}[theme=simple]
\begin{deptext}[column sep=.28cm]
I-ga \& children-o \& Kim-ni \& house-o \& let \& help \& paint \\
\end{deptext}

\depedge[edge style={SkyBlue, ultra thick, out=30, in=150}]{2}{5}{}
\depedge[edge style={Yellow, ultra thick, out=30, in=150}]{3}{6}{}
\depedge[edge style={Rhodamine, ultra thick, out=30, in=150}]{4}{7}{}

\end{dependency}

\caption{Basic CSD.}
\end{subfigure}

\\[.35em]

%================================================
% Rel
%================================================
\begin{subfigure}[t]{\textwidth}
\centering
\small
\begin{dependency}[theme=simple]
\begin{deptext}[column sep=.18cm]
I-ga \& $\lbrack$ \& I-ga \& met \& who \& $\rbrack$ \& children-o \& Kim-ni \& house-o \& let \& help \& paint \\
\end{deptext}

\depedge[edge style={SkyBlue, ultra thick, out=30, in=150}]{7}{10}{}
\depedge[edge style={Yellow, ultra thick, out=30, in=150}]{8}{11}{}
\depedge[edge style={Rhodamine, ultra thick, out=30, in=150}]{9}{12}{}

\depedge[edge style={black, thick, out=30, in=150}]{4}{7}{}

\end{dependency}

\caption{CSD + a single relative clause (\textsc{Rel}).}
\end{subfigure}

\\[.35em]

%================================================
% Conj-1
%================================================
\begin{subfigure}[t]{\textwidth}
\centering
\small
\begin{dependency}[theme=simple]
\begin{deptext}[column sep=.14cm]
I-ga \& $\lbrack$ \& I-ga \& met \& and \& I-ga \& liked \& who \& $\rbrack$ \& children-o \& Kim-ni \& house-o \& let \& help \& paint \\
\end{deptext}

\depedge[edge style={SkyBlue, ultra thick, out=30, in=150}]{10}{13}{}
\depedge[edge style={Yellow, ultra thick, out=30, in=150}]{11}{14}{}
\depedge[edge style={Rhodamine, ultra thick, out=30, in=150}]{12}{15}{}

\depedge[edge style={black, thick, out=30, in=150}]{4}{10}{}
\depedge[edge style={black, thick, out=30, in=150}]{7}{10}{}

\end{dependency}

\caption{CSD + conjoined relative clauses with one relativizer (\textsc{Conj-1}).}
\end{subfigure}

\\[.35em]

%================================================
% Conj-2
%================================================
\begin{subfigure}[t]{\textwidth}
\centering
\small
\begin{dependency}[theme=simple]
\begin{deptext}[column sep=.12cm]
I-ga \& $\lbrack$ \& I-ga \& met \& who \& and \& I-ga \& liked \& who \& $\rbrack$ \& children-o \& Kim-ni \& house-o \& let \& help \& paint \\
\end{deptext}

\depedge[edge style={SkyBlue, ultra thick, out=30, in=150}]{11}{14}{}
\depedge[edge style={Yellow, ultra thick, out=30, in=150}]{12}{15}{}
\depedge[edge style={Rhodamine, ultra thick, out=30, in=150}]{13}{16}{}

\depedge[edge style={black, thick, out=30, in=150}]{4}{11}{}
\depedge[edge style={black, thick, out=30, in=150}]{8}{11}{}

\end{dependency}

\caption{CSD + conjoined relative clauses with two relativizers (\textsc{Conj-2}).}
\end{subfigure}

\\[.35em]

%================================================
% Nested
%================================================
\begin{subfigure}[t]{\textwidth}
\centering
\small
\begin{dependency}[theme=simple]
\begin{deptext}[column sep=.12cm]
I-ga \& $\lbrack$ \& I-ga \& met \& who \& $\lbrack$ \& man-ga \& liked \& who \& $\rbrack$ \& $\rbrack$ \& children-o \& Kim-ni \& house-o \& let \& help \& paint \\
\end{deptext}

\depedge[edge style={SkyBlue, ultra thick, out=30, in=150}]{12}{15}{}
\depedge[edge style={Yellow, ultra thick, out=30, in=150}]{13}{16}{}
\depedge[edge style={Rhodamine, ultra thick, out=30, in=150}]{14}{17}{}

\depedge[edge style={black, thick, out=30, in=150}]{4}{7}{}
\depedge[edge style={black, thick, out=30, in=150}]{8}{12}{}

\end{dependency}

\caption{CSD + nested relative clauses (\textsc{Nested}).}
\end{subfigure}

\end{tabular}
\end{adjustbox}

\caption{
Partial dependency representations of CSDs with embedded relative clauses.
The examples are for SOV with prenominal relative clauses (\texttt{0XXX0}); the pattern will be, for example, fully mirrored for VOS with postnominal relative clauses (\texttt{1XXX1}).
Colored arcs indicate cross-serial noun--verb dependencies.
Black arcs indicate relative-clause attachment.
Relative clauses are bracketed.
}
\label{fig:extending_cross_serial_templates}

\end{figure*}

\begin{algorithm}[h]
\small
\caption{Cross-Serial Template Expansion}
\label{alg:crossserial-extension}
\begin{algorithmic}[1]

\Require Switch $s$, parser $P$, bounds $L_{\min}, L_{\max}$, set of substrings $\mathcal{S}$
\Ensure Extended template set $\mathcal{T}_{new}$

\State Initialize $\mathcal{T}_{new} \gets \emptyset$
\State Select $\mathcal{T}_{short}, \mathcal{T}_{other}$ based on $s$

\If{$s$ starts with ``0''}
    \State $\mathcal{T}_{new} \gets \mathcal{T}_{new} \cup \{\text{canonical SOV}\}$
\Else
    \State $\mathcal{T}_{new} \gets \mathcal{T}_{new} \cup \{\text{canonical VOS}\}$
\EndIf

\Statex

\Function{Expand}{$\mathcal{T}$}
\ForAll{$t \in \mathcal{T}$}
    \ForAll{substring $u \in \mathcal{S}$}
        \ForAll{insertion position $i$ in $t$}
            \State $t' \gets \text{Insert}(t,u,i)$
            \If{$L_{\min} \le |t'| \le L_{\max}$ \textbf{and} $P(t')$ valid}
                \State $\mathcal{T}_{new} \gets \mathcal{T}_{new} \cup \{t'\}$
            \EndIf
        \EndFor
    \EndFor
\EndFor
\EndFunction

\State \Call{Expand}{$\mathcal{T}_{short}$}
\State \Call{Expand}{$\mathcal{T}_{other}$}

\ForAll{$t_1, t_2 \in \mathcal{T}_{new} \cup \mathcal{T}_{short} \cup \mathcal{T}_{other}$}
    \State $t' \gets t_1 \;\texttt{conj}\; t_2$
    \If{$L_{\min} \le |t'| \le L_{\max}+2$ \textbf{and} $P(t')$ valid}
        \State $\mathcal{T}_{new} \gets \mathcal{T}_{new} \cup \{t'\}$
    \EndIf
\EndFor

\State Remove $\mathcal{T}_{medium} \cup \mathcal{T}_{short}$ from $\mathcal{T}_{new}$
\State Output $\mathcal{T}_{new}$

\end{algorithmic}
\end{algorithm}

\section{Artificial Language Datasets}

\subsection{Generalized Categorial Grammar}
\label{sec:generalized_categorial_grammar}

The ALs we use are based on generalized categorial grammar (GCG) \cite{wood2014categorial}. 
GCGs are an extension of categorial grammar (CG).
CG is made up of a lexicon, where each word is assigned a functor or basic category, and rules that determine how categories are combined syntactically and semantically.
The different categories are represented using slash notation, which indicates the position of the argument with respect to the resulting category, i.e., whether the argument is on the left or right of the result.
The 2 combinatory operations included in CG are forward (a) and backward (b) functional application:
\begin{enumerate}
    \item[(a)] $A/B \text{ } B \Rightarrow A$
    \item[(b)] $B \text{ } A$ \textbackslash{} $B \Rightarrow A$
\end{enumerate}
Forward and backward application are demonstrated below using the English transitive verb "met", which is the functor category (S\textbackslash{}NP)/NP.
\begin{center}
\deriv{3}{

\text{Bart} & \text{met} & \text{Lisa} \\
\uline{1} & \uline{1} & \uline{1}\\
\text{NP} & \text{(S\bs NP)/NP} & \text{NP}\\
 & \fapply{2} \\
 & \mc{2}{\text{S\bs NP}}\\
 \bapply{2} & \\
 \mc{2}{\text{S}} & \\
}
\end{center}

Extensions of CG, like combinatory categorial grammar (CCG) \cite{steedman1996surface} and GCG, introduce additional combination operations, such as forward (a) and backward (b) composition (\textbf{B}):
\begin{enumerate}
    \item[(a)] $A/B \text{ } B/C \Rightarrow A/C$
    \item[(b)] $B$\textbackslash{} $C$ \text{ } $A$\textbackslash{} $B \Rightarrow A$\textbackslash{} $C$
\end{enumerate}
Composition (\textbf{B}) is demonstrated
% in the derivation 
below.
\begin{center}
    \deriv{4}{
    \text{the children} & \text{in} & \text{the park} & \text{played}\\
    \uline{1} & \uline{1} & \uline{1} & \uline{1} \\
    \text{NP} & \text{(NP\bs NP)/NP} & \text{NP} & \text{S\bs NP}\\
     & \fapply{2} & \\
     & \mc{2}{\text{NP\bs NP}} & \\
     & \bcomp{3}\\
     & \mc{3}{\text{S\bs NP}}\\
     \bapply{4}\\
     \mc{4}{\text{S}}\\
    }
\end{center}

Coordination ($\Phi$) is another operation in extensions of CG, where 2 elements of the same category separated by a conjunction can be combined into a single element of the same category.
This is demonstrated in the example below:
\begin{center}
\deriv{5}{
\text{Bart} & \text{and} &  \text{Lisa} & \text{watched} & \text{television}\\
\uline{1} & \uline{1} & \uline{1} & \uline{1 }& \uline{1}\\
\text{NP} & \text{CONJ} & \text{NP} &  \text{(S\bs NP)/NP} & \text{NP}\\
\conj{3} & & \\
\mc{3}{\text{NP}}&  \\
 & & &  \fapply{2} \\
 & & & \mc{2}{\text{S\bs NP}}\\
 & \bapply{4} \\
 & \mc{4}{\text{S}} \\
 }
\end{center}

CCG includes, type raising, which we do not include in our grammar, and instead include generalized cyclic permutation as defined by \citet{briscoe1997co,briscoe2000grammatical}.
Permutation (P) can be formally defined as:
$$(A|B_{1})...|B_{n} \Rightarrow (A|B_{n})|B_{1}$$
Permutation (P) is shown 
below:
\begin{center}
    \deriv{4}{
    \text{Bart} & \text{whom} & \text{Lisa} & \text{met}\\
    \uline{1} & \uline{1} &\uline{1} &\uline{1} \\
    \text{NP} & \text{(NP\bs NP)/(S/NP)} & \text{NP} & \text{(S\bs NP)/NP} \\
     & & & \permute{1} \\
     & & & \mc{1}{\text{(S/NP)\bs NP}} \\
     & & \bapply{2} \\
     & & \mc{2}{\text{S/NP}} \\
     & \bapply{3}\\
     
     & \mc{3}{\text{NP\bs NP}}\\
     \bapply{4} \\
     \mc{4}{\text{NP}} \\
    }
\end{center}
\subsection{GCG-based ALs}\label{sec:AL_details_original}
In our GCG, the following operations are used:
\begin{itemize}
    \item Application (forward and backward)
    \item Composition (forward and backward)
    \item Coordination
    \item Permutation
\end{itemize}

Table~\ref{table:category_descriptions} shows how slash directionalities in our lexical categories are controlled by word order parameters, which are described in Table~\ref{tab:wo_params}.

We use a modified version of the NLTK \cite{DBLP:books/daglib/0022921} CCGChartParser.
In this modified version of the NLTK CCGChartParser, type raising is disabled, and the permutation operation defined by \citet{briscoe1997co,briscoe2000grammatical} is implemented. 
Crossed composition is allowed in the modified parser, enabling it to parse CSD constructions and their natural extensions (CSD with relative clause, etc.).
This parser is used to parse the newly introduced templates and determine their validity. 

We
follow the methodology of ~\citet{el2025word}
to generate a set of ALs with different word orders, but the same word forms and thus tokenization across all the ALs. We make some modifications to align and extend the generated ALs and associated test sets with our research questions.
\citet{el2025word} define a parametrized generalized categorial grammar (GCG) and generate multiple variations of the AL corpora with different word orders by manipulating these parameters.
Table~\ref{table:category_descriptions} shows the lexical syntactic categories adopted in our ALs. 
These are combined to make a sentence (S), following the general combinatory syntactic rules of application, composition and permutation
(see §\ref{sec:AL_details_original}).
Note that we add crossed composition to the GCG parser, enabling it to parse CSD constructions and their natural extensions (CSD with relative clause, etc.).

AL corpora are created by filtering sentences generated from templates defining sequences of lexical categories using a GCG parser with language-specific sets of lexical syntactic categories.
Note that all our ALs have case markers, i.e., \textit{ga} for subjective and \textit{o} for accusative case, which will be used in example sentences. Lexical items are randomly selected to create sentences from templates based on their assignment to a single lexical category (i.e. there is no lexical syntactic ambiguity in our ALs).

For each AL, data are split into 70K \textsc{Short} (3-8 tokens) sentences in training data, 10K held-out \textsc{Short} validation data, 20K held-out \textsc{Short} test data, 20K \textsc{Medium} (9-10 tokens) test data, and 20K \textsc{Long} (10-20 tokens) test data.
Note that the simplest CSD construction with 3-NP 2-VP  (e.g., \textit{We ga \textcolor{red}{Hans ga} \textcolor{blue}{house o} \textcolor{red}{helped} \textcolor{blue}{paint}}), is in the \textsc{Short} training set.
More complex ones, such as 4-NP 3-VP or CSD with some modifiers, are in the longer test data.
The modifications made to extend CSD constructions for the \textsc{Medium} and \textsc{Long} test sets are detailed in Algorithm~\ref{alg:crossserial-extension} creating additional templates.

\subsection{Modifications to the ALs}
We make four modifications to the ALs training and test datasets from~\citet{el2025word}; otherwise, the method of creating ALs is identical.

First, to 
include
CSD constructions,
we add a dative case marker (\textit{ni}) and several causative verb types (see Table~\ref{table:category_descriptions}). 
We also include relativizers that can modify NPs with accusative and dative object case markers.

Second, we refined the 
creation of
\textsc{Long} test data. 
While the previous study just concatenated two templates to create unseen examples, we added sentences from extended short templates (see Algorithm~\ref{alg:crossserial-extension}) and 4-NP 3-VP versions of CSD constructions to an extended version of the long test dataset.

Third, we limited the base word order variations to SOV and VOS; that is, the parameters determining subject, object, and verb are merged into a single binary parameter (0 for SOV, 1 for VOS).
This is because CSD constructions are only attested in SOV and VOS languages~\cite{shieber1985evidence,maclachlan2002cross}, and therefore, it becomes less obvious how to emulate CSD constructions for other base word orders, and targeted evaluations cannot be designed in a manner informed by attested languages.
Other parameters, such as adjective word order, are systematically controlled, resulting in 32 ALs and 5 word order parameters ($2^5=32$).
Word order configuration is denoted by a 5-digit binary code (Table~\ref{tab:wo_params}); for example, Japanese-like fully head-final word order is \texttt{00000}.

Fourth, we introduce new evaluation sets targeted only at CSD constructions, which are explained in the next section.

\subsection{CSD-Focused Generalization Test Sets}
To evaluate whether (S)LMs generalize to complex constructions, specifically the combination of CSDs and other syntactically complex ones, e.g., object relative clauses, we create four versions of evaluation data with specific constructions.
Figure~\ref{fig:extending_cross_serial_templates} illustrates the four constructions targeted in this study.
All the templates expand the basic 3-NP 2-VP and 4-NP 3-VP CSD constructions.
\textsc{Rel} version introduces an object relative clause to one of the NPs in the base construction.
\textsc{Conj-1} and \textsc{Conj-2} introduce a coordinate structure within the relative clause on top of the \textsc{Rel} version with one or two relativizers, respectively.
\textsc{Nested} introduces one additional nested relative clause into the \textsc{Rel} version.
In all versions, lexical items of the appropriate categories are randomly sampled. Each test set consists of 20K sentences.
\footnote{Datasets and model scripts are in \url{https://github.com/kuribayashi4/typological-alignment-stackLM}}

\begin{figure*}[t]
    \includegraphics[width=0.98\textwidth]{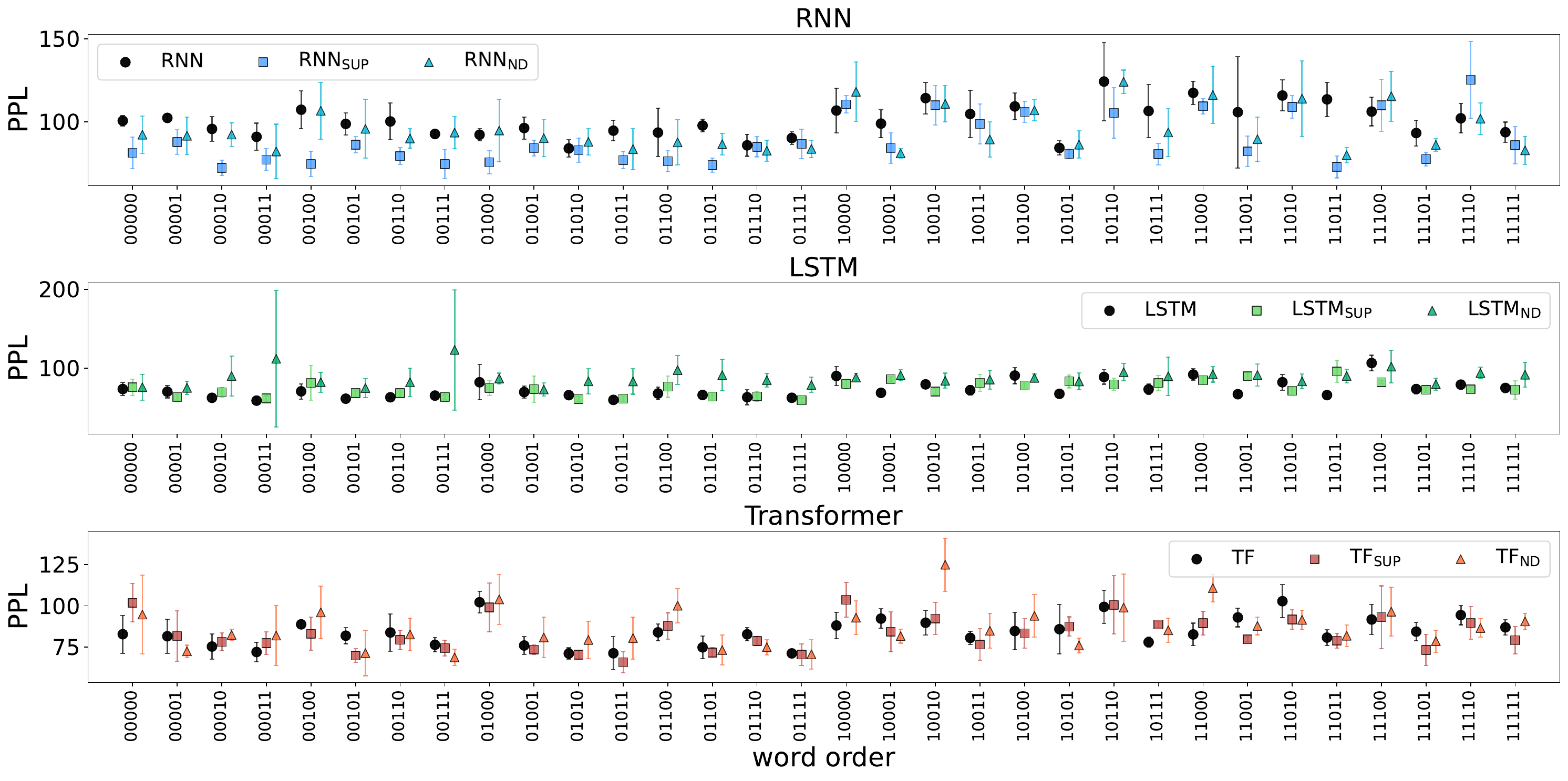}
    \caption{PPL distribution on the \textsc{Medium} test set (average$\pm$standard deviation). The bar shows the standard deviation among five runs with different random seeds. Note that all training was conducted and stopped under the same early-stopping setting, and the high variance indicates training instability across different random seeds (see \cref{sec:discussion}).}
    \label{fig:ppl_dist}
\end{figure*}

\section{Experimental Settings}

\subsection{Models}
We use the model variations and implementations of~\citet{dusell2025bearingsyntacticfruitstackaugmented}. 
As the stack module can be combined with different base architectures, we explore three base architectures: RNN, LSTM, and Transformer.
Two stack implementations of SUP and ND are examined, resulting in 6 SLM variants: \{RNN, LSTM, Transformer\}$\times$\{SUP, ND\}.
All the RNN and LSTM variants have 3 layers, and Transformers have 5 layers.
For RNN and LSTM SLMs, as proposed in~\citet{dusell2025bearingsyntacticfruitstackaugmented}, their multi-layer implementations are adopted, and we also add a short-circuit connection from the stack-top to the final layer by default to mitigate the time-lag between stack updating and reading.
For Transformer SLMs, we examine the two-stack-layer versions adopted in~\citet{dusell2025bearingsyntacticfruitstackaugmented} with the potential expressivity to recognize cross-serial constructions, and sinusoidal positional encoding is used.
See Appendix~\ref{app:ablation} for the detailed ablations of short-circuit connection and one-stack-layer Transformer SLMs.   
SUP stacks use a 50-dim stack vector, and ND use a 45-dim vector, where a 5-dim vector is assigned to nine different action histories (3 states$\times$3 stack tops).
A dropout rate of 0.1 is applied for all the models.

To facilitate fair comparison of models,  all the (S)LMs are set to have approximately 200K parameter counts.
We ran a model-data-dependent hyperparameter search to determine minibatch size and initial learning, instead of fixing the hyperparameters across all settings.
Specifically, for each model, we run ten hyperparameter configurations and select the one with the lowest cross-entropy loss on the validation set after 10 epochs.
We adopted early stopping after no improvement over three consecutive epochs, resulting in an average of 18 training epochs.

\subsection{Evaluation}
For each combination of models and word-order configurations, we train a model on the \textsc{Short} training data from scratch and evaluate its perplexity (PPL) on six different evaluation datasets.
We run five random seeds for each setting and report the average PPL across them.
For the typological alignment between PPLs and word-order frequency, we adopt the word-order counts from the World Atlas of Language Structures (WALS) \cite{wals}, following existing studies.

\begin{table*}[t]
\centering
\scriptsize
\tabcolsep=0.15cm
\begin{tabular}{lrrrrrrrr}
\toprule
Model & Short & Medium & Long & Rel & Conj-1 & Conj-2 & Nested & Avg. \\
\midrule
RNN & 33.40 $\pm$ 0.14 & 100.69 $\pm$ 13.91 & 345.68 $\pm$ 510.02 & 376.50 $\pm$ 223.62 & 552.94 $\pm$ 483.43 & 731.62 $\pm$ 1033.28 & 538.97 $\pm$ 574.27 & 382.83 \\
RNN$_{\text{SUP}}$ & \textbf{\underline{33.32}} $\pm$ 0.18 & \textbf{\underline{88.29}} $\pm$ 16.40 & \textbf{\underline{249.02}} $\pm$ 75.80 & \textbf{\underline{299.37}} $\pm$ 97.15 & \textbf{\underline{404.95}} $\pm$ 126.49 & \textbf{\underline{485.74}} $\pm$ 160.56 & \textbf{\underline{437.99}} $\pm$ 171.39 & \textbf{\underline{285.53}} \\
RNN$_{\text{ND}}$ & 33.43 $\pm$ 0.13 & \textbf{94.98} $\pm$ 16.35 & \textbf{333.45} $\pm$ 134.35 & \textbf{328.11} $\pm$ 101.49 & \textbf{497.27} $\pm$ 165.78 & \textbf{615.10} $\pm$ 241.20 & \textbf{504.19} $\pm$ 245.97 & \textbf{343.79} \\
\cmidrule(lr){1-1} \cmidrule(lr){2-8} \cmidrule(lr){9-9}
LSTM & 33.49 $\pm$ 0.22 & \underline{72.72} $\pm$ 13.04 & \underline{313.71} $\pm$ 86.15 & \underline{277.15} $\pm$ 100.83 & \underline{501.66} $\pm$ 208.33 & \underline{688.17} $\pm$ 305.38 & \underline{522.85} $\pm$ 257.08 & \underline{344.25} \\
LSTM$_{\text{SUP}}$ & 33.49 $\pm$ 0.20 & 73.80 $\pm$ 11.75 & 415.24 $\pm$ 190.50 & 364.23 $\pm$ 154.01 & 661.19 $\pm$ 298.72 & 1057.02 $\pm$ 601.37 & 772.05 $\pm$ 377.53 & 482.43 \\
LSTM$_{\text{ND}}$ & 34.52 $\pm$ 7.51 & 88.03 $\pm$ 24.14 & 516.53 $\pm$ 226.21 & 422.71 $\pm$ 188.28 & 773.97 $\pm$ 388.83 & 1159.86 $\pm$ 664.59 & 882.36 $\pm$ 526.66 & 554.00 \\
\cmidrule(lr){1-1} \cmidrule(lr){2-8} \cmidrule(lr){9-9}
TF & 33.32 $\pm$ 0.17 & 84.09 $\pm$ 10.85 & 289.18 $\pm$ 76.28 & 286.99 $\pm$ 70.01 & \underline{490.09} $\pm$ 186.07 & \underline{606.43} $\pm$ 283.38 & 419.58 $\pm$ 134.27 & \underline{315.67} \\
TF$_{\text{SUP}}$ & 33.32 $\pm$ 0.21 & \textbf{\underline{82.95}} $\pm$ 12.81 & 291.94 $\pm$ 79.25 & 290.00 $\pm$ 83.24 & 508.64 $\pm$ 218.43 & 620.30 $\pm$ 301.98 & 423.25 $\pm$ 156.09 & 321.49 \\
TF$_{\text{ND}}$ & \textbf{\underline{33.32}} $\pm$ 0.20 & 86.75 $\pm$ 15.94 & 358.00 $\pm$ 101.45 & 307.90 $\pm$ 89.49 & 537.66 $\pm$ 214.94 & 654.70 $\pm$ 296.13 & 449.55 $\pm$ 156.11 & 346.84 \\
\bottomrule
\end{tabular}
\caption{PPL for each model and evaluation data (average$\pm$ standard deviation). The standard deviation here indicates how the average PPL varies among 32 languages. PPL values obtained from SLMs, which are better than those of the corresponding base model, are shown in \textbf{bold}. The best PPLs among the LMs with the same base architecture are shown with an \underline{underline}.}
\label{tab:ppl}
\end{table*}

\section{Experiments}

We train (S)LMs on the short training sets for each AL, and measure the models' perplexity on the corresponding test sets.
This gives the PPL distribution over 32 ALs with different word order configurations.
Taking the \textsc{Medium} test set as an example, Figure~\ref{fig:ppl_dist} shows the PPL distributions for each model.
The X-axis shows word order variations, the 5-digit codes represent the word order parameter settings (see Table~\ref{tab:wo_params}).

\begin{table}[t]
\centering
\small
\tabcolsep=0.001cm
\begin{tabular}{p{1.05cm}rrrrrrrr}
\toprule
Model & Short & Med. & Long & Rel & Conj1 & Conj2 & Nest \\
\cmidrule(lr){1-1} \cmidrule(lr){2-8}
RNN & 0.57\phantom{$^\dagger$} & $-$0.66$^\dagger$ & $-$0.57$^\dagger$ & $-$0.31$^\dagger$ & $-$0.48$^\dagger$ & $-$0.59$^\dagger$ & $-$0.25\phantom{$^\dagger$} \\
RNN$_{\text{SUP}}$ & $-$0.24\phantom{$^\dagger$} & $-$0.58$^\dagger$ & $-$0.43$^\dagger$ & $-$0.60$^\dagger$ & $-$0.62$^\dagger$ & $-$0.45$^\dagger$ & $-$0.39$^\dagger$ \\
RNN$_{\text{ND}}$ & 0.36\phantom{$^\dagger$} & $-$0.40$^\dagger$ & $-$0.24\phantom{$^\dagger$} & $-$0.61$^\dagger$ & $-$0.26\phantom{$^\dagger$} & $-$0.12\phantom{$^\dagger$} & $-$0.36$^\dagger$ \\
\cmidrule(lr){1-1} \cmidrule(lr){2-8}
LSTM & 0.43\phantom{$^\dagger$} & $-$0.56$^\dagger$ & $-$0.69$^\dagger$ & $-$0.57$^\dagger$ & $-$0.26\phantom{$^\dagger$} & $-$0.28\phantom{$^\dagger$} & $-$0.38$^\dagger$ \\
LSTM$_{\text{SUP}}$ & 0.54\phantom{$^\dagger$} & $-$0.60$^\dagger$ & $-$0.66$^\dagger$ & $-$0.49$^\dagger$ & $-$0.48$^\dagger$ & $-$0.56$^\dagger$ & $-$0.55$^\dagger$ \\
LSTM$_{\text{ND}}$ & 0.59\phantom{$^\dagger$} & $-$0.25\phantom{$^\dagger$} & 0.17\phantom{$^\dagger$} & $-$0.24\phantom{$^\dagger$} & 0.10\phantom{$^\dagger$} & 0.20\phantom{$^\dagger$} & 0.12\phantom{$^\dagger$} \\
\cmidrule(lr){1-1} \cmidrule(lr){2-8}
TF & 0.02\phantom{$^\dagger$} & $-$0.58$^\dagger$ & 0.12\phantom{$^\dagger$} & 0.35\phantom{$^\dagger$} & 0.58\phantom{$^\dagger$} & 0.49\phantom{$^\dagger$} & 0.36\phantom{$^\dagger$} \\
TF$_{\text{SUP}}$ & 0.01\phantom{$^\dagger$} & $-$0.44$^\dagger$ & 0.04\phantom{$^\dagger$} & 0.40\phantom{$^\dagger$} & 0.62\phantom{$^\dagger$} & 0.50\phantom{$^\dagger$} & 0.22\phantom{$^\dagger$} \\
TF$_{\text{ND}}$ & $-$0.21\phantom{$^\dagger$} & $-$0.36$^\dagger$ & 0.27\phantom{$^\dagger$} & 0.19\phantom{$^\dagger$} & 0.59\phantom{$^\dagger$} & 0.55\phantom{$^\dagger$} & 0.31\phantom{$^\dagger$} \\
\bottomrule
\end{tabular}
\caption{Spearman Correlation to Word Order Frequency by Model and Test/Evaluation Type. $\dagger$ indicates a significantly negative correlation (p < 0.05).}
\label{tab:correlation}
\end{table}

\subsection{Averaged PPLs}
\label{subsec:ppl_results}
We first analyze the general architectural effects over diverse word orders.
Table~\ref{tab:ppl} shows the PPL averaged over 32 languages for each model and evaluation set.
These results show that stack modules do not always improve PPLs.
One notable result is the superiority of the combination of RNN and SUP stack, which yields consistently better PPLs than the base RNN across all the evaluation sets. This model also showed good generalization in other extant research~\cite{dusell2025bearingsyntacticfruitstackaugmented}.
In other words, at least under fixed training resources, SLMs with powerful context access (Transformer) and expressive stack implementations (nondeterministic stack) are not the best architectures for achieving good generalization.
This also shows that, in contrast to the theoretical analysis~\cite{DBLP:conf/iclr/DuSell023}, SLMs, even with nondeterministic stacks, do not show notable generalization ability when tested on more realistic ALs.

\subsection{Typological Alignment}
\label{subsec:typology_results}
We then evaluate which model's learning bias is more aligned with typological distributions.
One straightforward analysis is to compute the correlation between the typological frequency of the word order and the perplexity for each model and test set.
A negative correlation indicates that the model yields better (lower) PPLs for more common word order patterns, i.e., its learning bias is typologically aligned.
We then analyze how this correlation changes across different evaluation settings, especially for any advantage of stack representations.
Table~\ref{tab:correlation} shows the Spearman correlation coefficients between the typological frequency of the word order and the perplexity of the model on each evaluation setting.

First, we observe that typological alignment is more enhanced when the model's out-of-distribution generalization is targeted, rather than in-distribution fit, which is consistent with existing studies.
When the alignment is evaluated in targeted evaluation data with
CSDs,
the alignment of non-Transformer LMs with superposition stack representations is typically better than that of Transformer-based SLMs.
We statistically analyze these tendencies in the next section.

\begin{table}[t]
\centering
\small
\begin{tabular}{lrr}
\toprule
Feature & Coefficient & Significance \\
\midrule
$\beta_0$ & 0.0371 &  \\
$\beta_F$ & $-$0.0651 & * \\
$\beta_{M(\text{LSTM})}$ & $-$0.5231 & *** \\
$\beta_{M(\text{TF})}$ & $-$0.0628 &  \\
$\beta_{S(\text{Sup})}$ & $-$0.0213 &  \\
$\beta_{S(\text{Nd})}$ & $-$0.0790 &  \\
$\beta_{M(\text{LSTM}):S(\text{SUP})}$  & 0.6574 & *** \\
$\beta_{M(\text{TF}):S(\text{SUP})}$ & 0.3034 & *** \\
$\beta_{M(\text{LSTM}):S(\text{ND})}$ & 0.8247 & *** \\
$\beta_{M(\text{TF}):S(\text{ND})}$& 0.3614 & *** \\
\midrule
$\beta_{F:M(\text{LSTM})}$ & 0.0146 &  \\
$\beta_{F:M(\text{TF})}$ & 0.1039 & *** \\
$\beta_{F:S(\text{SUP})}$  & $-$0.0847 & * \\
$\beta_{F:S(\text{ND})}$ & $-$0.0146 &  \\

\bottomrule
\end{tabular}
\caption{Fixed effects from the regression model predicting perplexity. 
The word order digit effects ($\boldsymbol{\beta}_d$, $\boldsymbol{\beta}_{d:M}$, $\boldsymbol{\beta}_{d:S}$)  are excluded from the Table.
Coefficients for each parameter $\beta_{\text{effect}}$ are shown. Significance markers denote: *** p<0.001, ** p<0.01, * p<0.05. }
\label{tab:regression_coefficients}
\end{table}

\subsection{Regression Analysis}
Using Statsmodels \cite{seabold2010statsmodels}, we fit a regression model to predict the PPL for each configuration to identify which factor explains the relative success of generalization, following~\citet{ning-etal-2026-effect}.

\paragraph{Notations}
Let the evaluation setting be indexed with $i$, which can be mapped to several attributes:
\begin{itemize}
    \item Obtained perplexity $\text{PPL}_i$
    \item Word order $\mathbf{d}_i \in \{0, 1\}^5$ that determines the BASE, COMP, PP, ADJ, and REL word order parameters
    \item Typological frequency of corresponding word order $F_i$
    \item Base model architecture \\ $\mathbf{M}_i \in \{\text{RNN}, \text{LSTM}, \text{Transformer}\}$
    \item Stack representation \\ $\mathbf{S}_i \in \{\text{no}, \text{SUP}, \text{ND}\}$
    \item Test data for the evaluation \\ $\mathbf{T}_i \in \{\text{S}, \text{M}, \text{L}, \text{Rel}, \text{Conj-1}, \text{Conj-2}, \text{Nested}\}$
\end{itemize}

\paragraph{Regression model}
We fit a regression model to predict the perplexity $\text{PPL}_i$ from a setting $i$ with word order $\mathbf{d}_i$, its frequency $F_i$, the model architecture $\mathbf{M}_i$, the stack representation $\mathbf{S}_i$, and their interactions:
\begin{equation*}
    \begin{aligned}
    \text{PPL}_i
     & = \beta_0  + \beta_F F_i +
     \boldsymbol{\beta}_d^\top \mathbf{d}_i  + \boldsymbol{\beta}_M^\top \mathbf{M}_i + \boldsymbol{\beta}_S^\top \mathbf{S}_i \\
     & \quad + \boldsymbol{\beta}_{F:M}^\top (F_i \circ \mathbf{M}_i) + \boldsymbol{\beta}_{d:M}^\top (\mathbf{d}_i \circ \mathbf{M}_i) \\
    & \quad  + \boldsymbol{\beta}_{F:S}^\top (F_i \circ \mathbf{S}_i) + \boldsymbol{\beta}_{d:M}^\top (\mathbf{d}_i \circ \mathbf{S}_i) \\
    & \quad + \boldsymbol{\beta}_{M:S}^\top (\mathbf{M}_i \circ \mathbf{S}_i) + \epsilon_i\;\;\mathrm{,} \\
\end{aligned}
\end{equation*}
\noindent
We train a single linear regression model to predict the PPLs for all the evaluation settings.
Note that each PPL value is an average of five different runs in each setting.
We use all the results, including the variations examined in Appendix~\ref{app:ablation}, in the regression modeling.

\paragraph{Results}
Table~\ref{tab:regression_coefficients} shows the coefficients of the regression model.
Note that RNN and no-stack settings are dummy-coded for $M$ and $S$, respectively.
From the perspective of typological alignment, interaction terms between typological frequency and architectural factors  $\boldsymbol{\beta}_{F:M}$ and $\boldsymbol{\beta}_{F:S}$ should be analyzed.
Specifically, these coefficients become negative if the architectural factors enhance the typological alignment, because a negative $\boldsymbol{\beta}_{F:S}$ indicates that the perplexity decreases more steeply as the typological frequency increases when the architectural factor is present.
A statistically significant negative coefficient is obtained for $\beta_{F:S(\text{Sup})}$, namely superposition stack representation.
Thus, we find the general advantage of the superposition stack representation in enhancing LMs' inductive bias to align with the attested typological distribution.

\section{Discussion}
\label{sec:discussion}
Despite the theoretical expressive power of ND stacks or powerful access to context of Transformers, the model achieving the best PPLs and typological alignment was the somewhat constrained RNN$_{\text{SUP}}$ model.
One interpretation is that the inductive bias of RNN$_{\text{SUP}}$ effectively regularizes learning and guides successful generalization.
This aligns with the well-known less-is-more hypothesis in human language acquisition \cite{newport1990maturational}, which proposes that cognitive limitations can aid language learning.
A recent study~\cite{mita-etal-2025-developmentally} shows that limited cognitive resources, i.e., working memory, often improve LMs' grammar acquisition empirically, and our study corroborates these findings from the distinct angles of AL learning and typological alignment. However, the resource limitation adopted in our study is not dynamically relaxed during training.
Our finding is also consistent with a related study~\cite{DBLP:conf/acl/KuribayashiUYOB24} where working-memory-limited syntactic LMs showed better typological alignment.

However, another reason for the differences across model architectures might be training stability.
As suggested in Figure~\ref{fig:ppl_dist}, models sometimes exhibit substantial instability across the five seeds within the same data and model combination. 
Note that we ensured all training was properly stopped via early stopping, so the seed variance reflects the model's inherent instability in training.
We confirmed that, for example, ND SLMs generally exhibit greater instability than SUP SLMs (see Appendix~\ref{app:seed_variations}); thus, the misalignment of ND SLMs could also be due to noise arising from unstable convergence, possibly because of their more complex model architectures.
From this perspective, while ND stacks may potentially be effective, we do not yet have an implementation that is robust enough to demonstrate this statistically.

\section{Conclusion}
In this paper, we evaluate RNN, LSTM, and Transformer LM learning of ALs with differing word order configurations focusing on generalization for test data containing cross-serial constructions.
We investigate the effect of augmenting our LMs with a stack on their inductive biases and their ability to learn and generalize.
We also investigate the alignment of LM learning with the attested typological distribution of word order configurations, and we find that the models with limited working memory, e.g., RNN$_{\text{SUP}}$ SLM model both performed most effectively, consistent with the less-is-more hypothesis based on human working memory limitations, and are more typologically aligned with the commonly occurring word order configurations in attested languages. 

\section*{Limitations}
The data used in this paper are artificial, where the sentences for all the ALs consist mostly of `English' word forms allowing us to abstract away from tokenization differences.
While using artificial languages provides us with an opportunity to conduct more controlled and targeted experiments, it also simplifies or excludes many other features that are found in natural languages. For instance, our current ALs exclude lexical category ambiguity --- no word form is associated with both a verb or noun for example, and no word form is both an intransitive and  transitive verb. Similarly, nearly all structural attachment ambiguity introduced by the combinatory rules is excluded. Further experiments should explore the effect of adding such ambiguity as well as explore a wider range of construction types.

Our results currently show that the LMs that incorporate the more expressive ND stack exhibit a higher variance in training and performance than the less expressive models like the vanilla models or the LMs with SUP stacks. 
We would like to better understand the reason for this behavior and training instability in the future.
Similarly, we would like to understand why the vanilla LSTM models learn cross-serial dependencies more effectively than stack-augmented LSTMs. Further experiments should also explore the effect of hyperparameter settings on the results.
Additionally, our results are based on empirical evaluation. 
To better understand the extent of SLM learning, we plan to conduct a more theoretical study in the future. 

\section*{Ethical Statement}
The datasets used in this paper are all artificial and automatically generated.
We have no ethical concerns or concerns for potential risks with the contents of this paper.

\section*{AI Writing/Coding Assistance Policy}
AI writing assistance systems like Grammarly and ChatGPT are used occasionally to correct grammatical and spelling errors and to adjust phrasing. 
Our use of writing assistants complies with the AI writing assistance policy described in ARR guidelines under category (a), assistance purely with the language of the paper. 

\section*{Acknowledgments}
This work was supported by JSPS KAKENHI Grant Numbers 23K16938 and 26H02511.

% Bibliography entries for the entire Anthology, followed by custom entries
%\bibliography{anthology,custom}
% Custom bibliography entries only
\bibliography{custom}

\appendix

\section{Information Relevant to Responsibility Checklist}
\label{sec:appendix_checklist}
\subsection{Model Details}
\label{app:model}
We follow the model configurations in~\citet{dusell2025bearingsyntacticfruitstackaugmented}.
The total parameter number is adjusted to be around 200K in all the models.
All the Transformer variants have five layers with four attention heads in each layer.
All the LSTM and RNN variants have three layers.
The dropout rate is set to 0.1.
The number of tokens in minibatch [512, 2048] and the initial learning rate [$10^{-5}$, $10^{-3}$] are determined by hyperparameter tuning, following~\citet{dusell2025bearingsyntacticfruitstackaugmented}.
The runs for hyperparameter search continues for 5 epochs, and the best hyperparameter combination is selected among 10 different runs.

The \textsc{Short} training, validation, and test data consist of $\sim$70K, $\sim$10K, and $\sim$20K sentences.
The other test data, including \textsc{Medium}, \textsc{Long}, Rel, Conj-1, Conj-2, and Nested data, have $\sim$20K sentences in each set.

The model training was run on four NVIDIA RTX 6000 Ada GPUs and took approximately 140 GPU hours in total.
\subsection{Artifacts}
\label{app:artifact}
The artifacts used in this paper are shown in Table~\ref{tab:artifacts}.
Table~\ref{tbl:wals} specifies which sections of WALS \cite{wals} and Grambank \cite{grambank_dataset_zenodo_v1} were referred to in estimating the frequency of word order configuration.

\begin{table*}[ht!]
    \centering
    \small
    \begin{tabular}{p{5cm}l p{6cm}}
    \toprule
        Artifact & License & Usage \\
        \cmidrule(lr){1-1} \cmidrule(lr){2-2} \cmidrule(lr){3-3} 
        
        NLTK \cite{DBLP:books/daglib/0022921} 
        & Apache License 2.0 
        & to create ALs and targeted evaluation data\\
        
        Docker \cite{merkel2014docker} 
        & Apache 2.0 
        & to train SLMs\\
        
        \citet{el2025word} Datasets 
        & Creative Commons CC-BY 4.0 
        & training and testing SLMs\\
        
        Codes  from \citet{dusell2025bearingsyntacticfruitstackaugmented} 
        (\url{https://github.com/bdusell/bearing-syntactic-fruit/commit/ebe4ee8}) 
        & MIT License 
        & to implement models \\
        
        Rau (\url{https://github.com/bdusell/rau}) 
        & MIT Licence 
        & to implement models \\
        
        WALS \cite{wals} 
        & Creative Commons CC-BY 4.0 
        & to find word order statistics in NLs\\
        
        Grambank \cite{grambank_dataset_zenodo_v1} 
        & Creative Commons CC-BY 4.0 
        & to find word order statistics in NLs\\
        
    \bottomrule
    \end{tabular}
    \caption{Artifacts used in this paper.}
    \label{tab:artifacts}
\end{table*}

\begin{table*}[ht!]
    \centering
\fontsize{8}{9}\selectfont
\setlength{\tabcolsep}{2pt}
\begin{tabular}{ll}
\toprule
S & 82A Order of Subject and Verb~\cite{wals-82} \\
VP & 83A Order of Object and Verb~\cite{wals-83} \\
O & 81A Order of Subject, Object and Verb~\cite{wals-81} \\
\multirow{2}{*}{COMP} & Feature GB421: Is there a preposed complementizer in complements of verbs of thinking and/or knowing?~\cite{grambank_dataset_zenodo_v1} \\
& Feature GB422: Is there a postposed complementizer in complements of verbs of thinking and/or knowing?~\cite{grambank_dataset_zenodo_v1} \\
PP & 85A Order of Adposition and Noun Phrase~\cite{wals-85} \\
ADJ & 87A Order of Adjective and Noun~\cite{wals-87} \\
REL & 90A Order of Relative Clause and Noun~\cite{wals-90} \\
\bottomrule
\end{tabular}
\caption{WALS and Grambank chapters used in this paper.}
\label{tbl:wals}
\end{table*}

\section{SLM Variants and Ablations}
\label{app:ablation}
Following~\citet{dusell2025bearingsyntacticfruitstackaugmented}, 
we also examined RNN/LSTM SLMs without the shortcut connection from stack reading to the final layer representation.
In addition, we examined Transformer SLMs with one stack sublayer.
Tables~\ref{tab:ppl_ablation} and \ref{tab:correlation_app} show results.

\section{Seed variation}
\label{app:seed_variations}
Table~\ref{tab:seed_variance} shows the coefficient of variation (CV) scores ($=\frac{\text{standard deviation}}{\text{mean}}$) of PPLs among five runs with different random seeds.
Stack-based RNNs yield comparable or better stability than base RNNs. 
Otherwise, Stack-based LSTMs and Transformers tend to yield higher CVs, i.e., worse stability, than their respective base models.

\begin{table*}[t]
\centering
\scriptsize
\tabcolsep=0.1cm
\begin{tabular}{lrrrrrrrr}
\toprule
Model & Short & Medium & Long & Rel & Conj-1 & Conj-2 & Nested & Avg. \\
\midrule
RNN & 33.40 $\pm$ 0.14 & 100.69 $\pm$ 13.91 & 345.68 $\pm$ 510.02 & 376.50 $\pm$ 223.62 & 552.94 $\pm$ 483.43 & 731.62 $\pm$ 1033.28 & 538.97 $\pm$ 574.27 & 382.83 \\
RNN$_{\text{SUP}}$ (w/o SC) & 33.44 $\pm$ 0.18 & \textbf{100.43} $\pm$ 13.04 & \textbf{292.64} $\pm$ 88.90 & \textbf{352.21} $\pm$ 127.53 & \textbf{506.62} $\pm$ 189.70 & \textbf{623.84} $\pm$ 258.64 & \textbf{533.75} $\pm$ 249.39 & \textbf{348.99} \\
RNN$_{\text{ND}}$  (w/o SC) & 33.45 $\pm$ 0.14 & \textbf{100.40} $\pm$ 12.66 & \textbf{271.09} $\pm$ 86.73 & \textbf{324.98} $\pm$ 89.09 & \textbf{428.46} $\pm$ 117.36 & \textbf{492.40} $\pm$ 150.42 & \textbf{447.41} $\pm$ 147.36 & \textbf{299.74} \\
\cmidrule(lr){1-1} \cmidrule(lr){2-8} \cmidrule(lr){9-9}
LSTM & 33.49 $\pm$ 0.22 & \underline{72.72} $\pm$ 13.04 & \underline{313.71} $\pm$ 86.15 & \underline{277.15} $\pm$ 100.83 & \underline{501.66} $\pm$ 208.33 & \underline{688.17} $\pm$ 305.38 & \underline{522.85} $\pm$ 257.08 & \underline{344.25} \\
LSTM$_{\text{SUP}}$ (w/o SC) & \textbf{\underline{33.45}} $\pm$ 0.18 & 75.78 $\pm$ 12.06 & 374.29 $\pm$ 133.49 & 301.13 $\pm$ 112.04 & 574.42 $\pm$ 283.83 & 850.53 $\pm$ 507.61 & 615.52 $\pm$ 351.33 & 403.59 \\
LSTM$_{\text{ND}}$ (w/o SC) & \textbf{33.48} $\pm$ 0.21 & 79.94 $\pm$ 14.51 & 359.74 $\pm$ 105.87 & 307.96 $\pm$ 106.65 & 537.33 $\pm$ 202.30 & 753.00 $\pm$ 314.25 & 581.48 $\pm$ 258.67 & 378.99 \\
\cmidrule(lr){1-1} \cmidrule(lr){2-8} \cmidrule(lr){9-9}
TF & 33.32 $\pm$ 0.17 & 84.09 $\pm$ 10.85 & 289.18 $\pm$ 76.28 & 286.99 $\pm$ 70.01 & \underline{490.09} $\pm$ 186.07 & \underline{606.43} $\pm$ 283.38 & 419.58 $\pm$ 134.27 & \underline{315.67} \\
TF$_{\text{SUP-1layer}}$ & 33.35 $\pm$ 0.22 & \textbf{83.33} $\pm$ 13.63 & \textbf{\underline{287.92}} $\pm$ 73.39 & \textbf{\underline{284.92}} $\pm$ 70.66 & 496.15 $\pm$ 181.11 & 617.21 $\pm$ 277.99 & \textbf{\underline{417.67}} $\pm$ 137.18 & 317.22 \\
TF$_{\text{ND-1layer}}$ & 33.34 $\pm$ 0.19 & 84.33 $\pm$ 16.23 & 328.40 $\pm$ 97.89 & 300.09 $\pm$ 92.06 & 522.07 $\pm$ 201.65 & 644.57 $\pm$ 282.78 & 439.83 $\pm$ 163.47 & 336.09 \\
\bottomrule
\end{tabular}
\caption{PPL for each dataset and evaluation data (average$\pm$ standard deviation). The standard deviation here indicates how the average PPL varies among 32 languages. ``SC'' denotes the shortcut connection. PPL values obtained from SLMs, which are better than those of the corresponding base model, are shown in \textbf{bold}. The best PPLs among the LMs with the same base architecture are shown with an \underline{underline}.}
\label{tab:ppl_ablation}
\end{table*}

\begin{table*}[t]
\centering
\tabcolsep=0.2cm
\begin{tabular}{lrrrrrrrr}
\toprule
Model & Short & Medium & Long & Rel & Conj-1 & Conj-2 & Nested \\
\cmidrule(lr){1-1} \cmidrule(lr){2-8}
RNN & 0.57\phantom{$^\dagger$} & $-$0.66$^\dagger$ & $-$0.57$^\dagger$ & $-$0.31$^\dagger$ & $-$0.48$^\dagger$ & $-$0.59$^\dagger$ & $-$0.25\phantom{$^\dagger$} \\
RNN$_{\text{SUP}}$ (w/o shortcut) & 0.52\phantom{$^\dagger$} & 0.04\phantom{$^\dagger$} & $-$0.07\phantom{$^\dagger$} & 0.15\phantom{$^\dagger$} & 0.30\phantom{$^\dagger$} & 0.39\phantom{$^\dagger$} & 0.29\phantom{$^\dagger$} \\
RNN$_{\text{ND}}$ (w/o shortcut)  & 0.36\phantom{$^\dagger$} & 0.18\phantom{$^\dagger$} & $-$0.31$^\dagger$ & $-$0.05\phantom{$^\dagger$} & 0.03\phantom{$^\dagger$} & 0.22\phantom{$^\dagger$} & 0.20\phantom{$^\dagger$} \\
\cmidrule(lr){1-1} \cmidrule(lr){2-8}
LSTM & 0.43\phantom{$^\dagger$} & $-$0.56$^\dagger$ & $-$0.69$^\dagger$ & $-$0.57$^\dagger$ & $-$0.26\phantom{$^\dagger$} & $-$0.28\phantom{$^\dagger$} & $-$0.38$^\dagger$ \\
LSTM$_{\text{SUP}}$ (w/o shortcut)  & 0.55\phantom{$^\dagger$} & $-$0.65$^\dagger$ & $-$0.56$^\dagger$ & $-$0.68$^\dagger$ & $-$0.38$^\dagger$ & $-$0.34$^\dagger$ & $-$0.48$^\dagger$ \\
LSTM$_{\text{ND}}$ (w/o shortcut) & 0.61\phantom{$^\dagger$} & $-$0.53$^\dagger$ & $-$0.58$^\dagger$ & $-$0.67$^\dagger$ & $-$0.39$^\dagger$ & $-$0.33$^\dagger$ & $-$0.44$^\dagger$ \\
\cmidrule(lr){1-1} \cmidrule(lr){2-8}
TF & 0.02\phantom{$^\dagger$} & $-$0.58$^\dagger$ & 0.12\phantom{$^\dagger$} & 0.35\phantom{$^\dagger$} & 0.58\phantom{$^\dagger$} & 0.49\phantom{$^\dagger$} & 0.36\phantom{$^\dagger$} \\
TF$_{\text{SUP-1layer}}$ & 0.01\phantom{$^\dagger$} & $-$0.64$^\dagger$ & $-$0.42$^\dagger$ & 0.09\phantom{$^\dagger$} & 0.47\phantom{$^\dagger$} & 0.33\phantom{$^\dagger$} & 0.08\phantom{$^\dagger$} \\
TF$_{\text{ND-1layer}}$ & 0.03\phantom{$^\dagger$} & $-$0.43$^\dagger$ & $-$0.04\phantom{$^\dagger$} & 0.13\phantom{$^\dagger$} & 0.46\phantom{$^\dagger$} & 0.43\phantom{$^\dagger$} & 0.11\phantom{$^\dagger$} \\
\bottomrule
\end{tabular}
\caption{Spearman Correlation to Word Order Frequency by Model and Test/Evaluation Type. $\dagger$ indicates significantly negative correlation (p < 0.05).}
\label{tab:correlation_app}
\end{table*}

\begin{table*}[t]
\centering
\small
\begin{tabular}{lrrrrrrr}
\toprule
Model & Short & Medium & Long & Rel & Conj-1 & Conj-2  & Nested \\
\midrule
RNN & 0.003 & 0.089 & 0.442 & 0.258 & 0.379 & 0.489 & 0.325 \\
RNN$_{\mathrm{SUP}}$ (w/o shortcut) & 0.003 & 0.089 & 0.216 & 0.153 & 0.176 & 0.203 & 0.172 \\
RNN$_{\mathrm{SUP}}$ & 0.004 & 0.088 & 0.188 & 0.184 & 0.197 & 0.217 & 0.205 \\
RNN$_{\mathrm{ND}}$  (w/o shortcut) & 0.003 & 0.089 & 0.190 & 0.172 & 0.191 & 0.220 & 0.173 \\
RNN$_{\mathrm{ND}}$ & 0.003 & 0.114 & 0.276 & 0.239 & 0.278 & 0.318 & 0.316 \\
\midrule
LSTM & 0.004 & 0.089 & 0.178 & 0.231 & 0.249 & 0.276 & 0.281 \\
LSTM$_{\mathrm{SUP}}$ (w/o shortcut)& 0.003 & 0.091 & 0.206 & 0.217 & 0.271 & 0.294 & 0.285 \\
LSTM$_{\mathrm{SUP}}$ & 0.003 & 0.092 & 0.229 & 0.260 & 0.276 & 0.307 & 0.303 \\
LSTM$_{\mathrm{ND}}$  (w/o shortcut) & 0.004 & 0.115 & 0.197 & 0.222 & 0.238 & 0.267 & 0.269 \\
LSTM$_{\mathrm{ND}}$ & 0.036 & 0.180 & 0.356 & 0.365 & 0.411 & 0.475 & 0.481 \\
\midrule
TF & 0.004 & 0.081 & 0.208 & 0.170 & 0.207 & 0.221 & 0.195 \\
TF$_{\mathrm{SUP-1layer}}$ & 0.005 & 0.099 & 0.195 & 0.181 & 0.190 & 0.201 & 0.195 \\
TF$_{\mathrm{SUP}}$ & 0.005 & 0.093 & 0.232 & 0.190 & 0.225 & 0.239 & 0.197 \\
TF$_{\mathrm{ND-1layer}}$ & 0.005 & 0.109 & 0.235 & 0.204 & 0.234 & 0.258 & 0.223 \\
TF$_{\mathrm{ND}}$ & 0.005 & 0.115 & 0.251 & 0.210 & 0.245 & 0.263 & 0.234 \\
\bottomrule
\end{tabular}
\caption{Coefficient of variations of PPLs among five different random seeds for each model and evaluation data. The value is averaged over 32 languages in each setting.}
\label{tab:seed_variance}
\end{table*}

\end{document}